\documentclass[10pt,twocolumn,letterpaper]{article}

\usepackage{cvpr}
\usepackage{times}
\usepackage{epsfig}
\usepackage{graphicx}
\usepackage{amsmath}
\usepackage{amssymb}
\usepackage{subfigure}
\usepackage{mathrsfs}
\usepackage{bm}
\usepackage{multirow}

\usepackage[pagebackref=true,breaklinks=true,letterpaper=true,colorlinks,bookmarks=false]{hyperref}

\cvprfinalcopy 

\def\cvprPaperID{808} 

\ifcvprfinal\fi
\begin{document}

\title{BridgeNet: A Continuity-Aware Probabilistic Network for Age Estimation}

\author{Wanhua Li$^{1,2,3,4}$, Jiwen Lu$^{1,2,3,*}$, Jianjiang Feng$^{1,2,3}$, Chunjing Xu$^{4}$, Jie Zhou$^{1,2,3}$, Qi Tian$^{4}$\\
$^{1}$Department of Automation, Tsinghua University, China \\
$^{2}$State Key Lab of Intelligent Technologies and Systems, China \\
$^{3}$Beijing National Research Center for Information Science and Technology, China \\
$^{4}$Noah's Ark Lab, Huawei \\
{\tt\small li-wh17@mails.tsinghua.edu.cn \qquad \{lujiwen,jfeng,jzhou\}@tsinghua.edu.cn}\\
{\tt \small \{xuchunjing,tian.qi1\}@huawei.com}}

\maketitle
\thispagestyle{empty}
\newcommand\blfootnote[1]{%
\begingroup
\renewcommand\thefootnote{}\footnote{#1}%
\addtocounter{footnote}{-1}%
\endgroup
}
\blfootnote{$^{*}$   Corresponding Author}

\begin{abstract}
   Age estimation is an important yet very challenging problem in computer vision.
   Existing methods for age estimation
   usually apply a divide-and-conquer strategy to deal with heterogeneous data
   caused by the non-stationary aging process. However, the facial aging process
   is also a continuous process, and the continuity relationship between different
   components has not been effectively exploited.
   In this paper, we propose  BridgeNet for age estimation, which aims to
   mine the  continuous relation between age labels effectively. The proposed
   BridgeNet consists of local regressors and gating networks. Local regressors
   partition the data space into multiple overlapping subspaces to tackle
   heterogeneous data and gating networks learn continuity aware weights for the
   results of local regressors by employing the proposed bridge-tree structure, which
   introduces bridge connections into tree models to enforce the similarity
   between neighbor nodes. Moreover, these two components of BridgeNet can be jointly
   learned in an end-to-end way.
   We show experimental results on the MORPH II, FG-NET and Chalearn LAP 2015
   datasets and find that BridgeNet outperforms the state-of-the-art methods.
\end{abstract}

\section{Introduction}

Age estimation attempts to predict the real age value or age group
based on facial images,
which is an important task in computer vision
due to the broad applications such as visual surveillance \cite{6126248},
human-computer interaction \cite{Geng2006Learning}, social media \cite{Rothe2016Deep},
and face retrieval \cite{Lanitis2004Comparing}, \emph{etc}. Although this
problem has been extensively studied for many years, it is still very challenging
to estimate human age precisely from a single image.

%

\begin{figure}[t]
\begin{center}
   \includegraphics[width=1.0\linewidth]{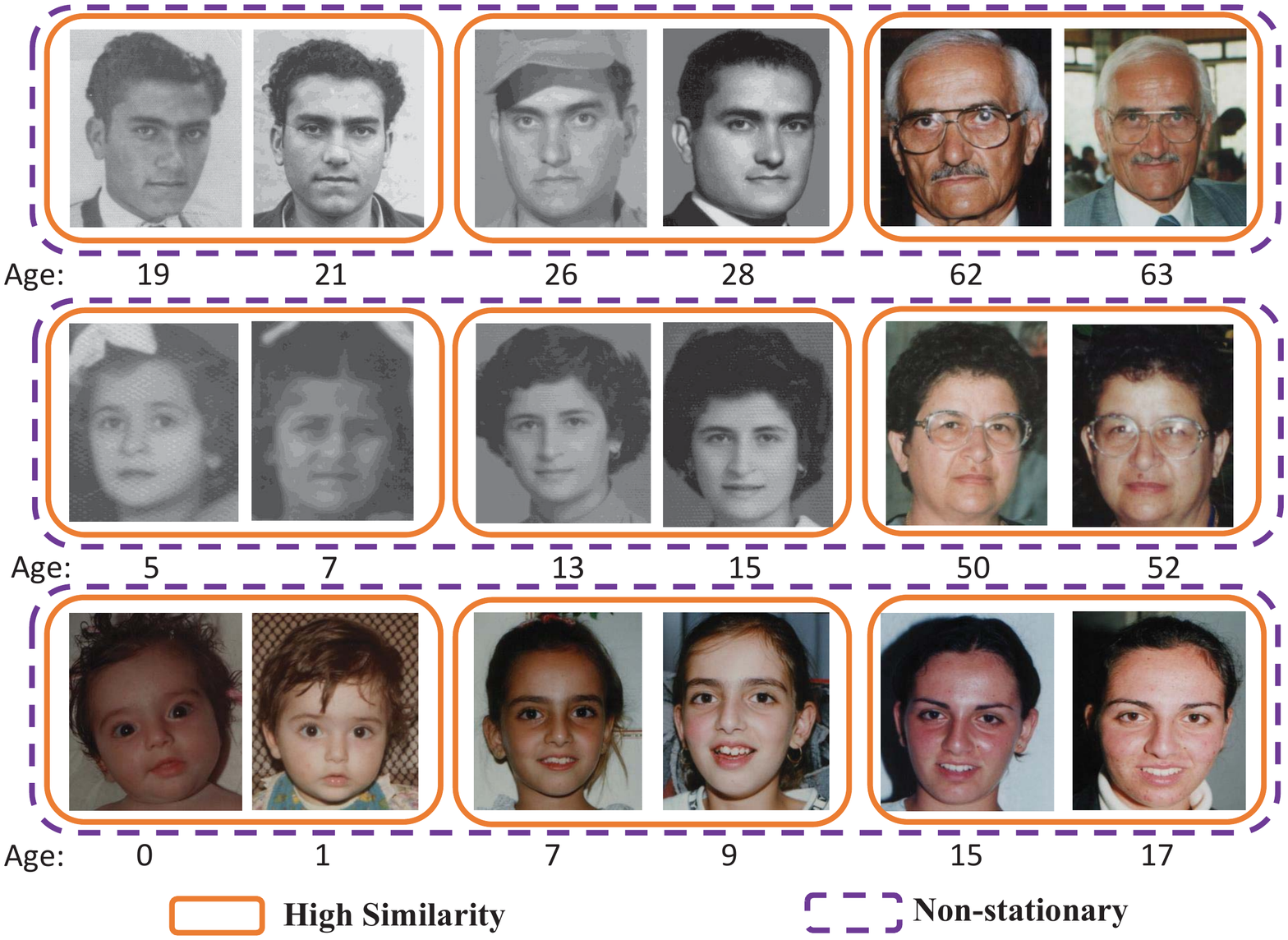}
\end{center}
   \caption{Facial images at different ages. The images of each row come from the same person.
    On the one hand, we can see the \textbf{non-stationary} property of aging patterns. The facial aging
    process is mainly reflected in the shape of the face during childhood and skin texture during
    adulthood. On the other hand, the facial images at adjacent ages show a very \textbf{high similarity}
    caused by the \textbf{continuous} aging process. }
\label{fig:movation}
\end{figure}


Age estimation can be cast as a regression problem by treating age labels as numerical values.
However, the human face matures in different
ways at different ages, e.g., bone growth in childhood and skin wrinkles in adulthood
 \cite{Ramanathan2009Age}. This \emph{non-stationary} aging process implies that the data of age estimation
is \emph{heterogeneous}. Thus many nonlinear regression approaches\cite{Guo2008Image,5206681} are inevitably
biased by the heterogeneous data distribution, and they are apt to overfit the training data
\cite{Chang2011Ordinal}.
Many efforts \cite{Shen_2018_CVPR,Rothe2016Deep,5206681,Niu2016Ordinal}
have been devoted to addressing this problem. Divide-and-conquer proves to be a good strategy to tackle
the heterogeneous data \cite{Huang2017Soft}, which divides the data space into multiple subspaces.
Huang \emph{et al.} use local regressors to learn homogeneous data partitions \cite{Huang2017Soft}.
Many ranking based methods transform the regression problem into a series of binary classification
subproblems \cite{Chen2017Using,Chang2011Ordinal}.
On the other hand, the facial aging process is also a \emph{continuous} process, that is
to say, human faces change gradually with age. Such a continuous process causes the
appearance of faces to be very similar at adjacent ages. For example, the facial appearance
will be very similar when you are 31 and 32. More examples are shown in Figure \ref{fig:movation}.
This similarity relationship caused by continuity
plays a dominant role at adjacent ages.
The same phenomenon can be found in adjacent local regressors or adjacent binary classification
subproblems, considering that we divide data by ages.
However, this relationship is not exploited in existing methods.

In this paper, we propose a continuity-aware probabilistic network, called BridgeNet,
to address the above challenges.
The proposed BridgeNet consists of local regressors and gating networks. The local
regressors partition the data space and gating networks provide continuity-aware weights.
The mixture of weighted regression results gives the final accurate estimation.
BridgeNet has many advantages. First, heterogeneous data are explicitly modeled by local
regressors as a divide-and-conquer approach.
Second, gating networks have a bridge-tree structure, which is proposed by introducing bridge
connections into tree models to enforce the similarity between
neighbor nodes on the same layer of bridge-tree.
Therefore, the gating networks can be aware of the continuity between local regressors.
Third,
the gating networks of BridgeNet use a probabilistic  soft decision instead of a hard decision, so
that the ensemble of local regressors can give a precise and robust estimation.
Fourth, we can jointly train  local regressors and  gating networks, and easily  integrate BridgeNet with
any deep neural networks into an end-to-end model.
We validate the proposed BridgeNet for age estimation on three  challenging datasets: MORPH Album II
\cite{Ricanek2006MORPH}, FG-NET \cite{Panis2016Overview}, and Chalearn LAP 2015 datasets \cite{Escalera2015ChaLearn},
and the experimental results demonstrate that our approach outperforms the state-of-the-art methods.
\section{Related Work}

\textbf{Age Estimation:}
Existing methods for age estimation can be grouped
into three categories: regression based methods, classification
 based methods, and ranking based methods \cite{Pan_2018_CVPR}.
 Regression based methods treat age labels as numerical values
 and utilize a regressor to regress the age.
  Guo \emph{et al.} introduced many regression based methods for
age estimation, such as SVR, PLS, and  CCA \cite{Guo2013Joint,Guo2011Simultaneous,5206681}. Zhang \emph{et al.}
proposed the multi-task warped Gaussian process \cite{Zhang2010Multi} to predict the age of face images.
However, these universal regressors suffer from
handling heterogeneous data. Hierarchical models \cite{Han2015Demographic} and group-specific regression
have shown promising results by dividing data by ages. Huang \emph{et al.} presented Soft-margin Mixture of
Regression to learn homogeneous partitions and learned a local regressor for each partition \cite{Huang2017Soft}.
But the continuity relationship between partitioned components is ignored in these methods.
Classification based methods usually treat different ages or age groups as independent class labels
\cite{5206681}. DEX \cite{Rothe2016Deep} cast age estimation as a classification problem with 101 categories.
Therefore, the costs of any type of classification error are the same, which can't
exploit the relations between age labels.
Recently, several researchers introduced ranking techniques to the
problem of age estimation. These methods usually utilize a series of simple binary classifiers  to determine
the rank of the age for a given input face image. Then the final age value can be obtained by combining
the results of these binary classification subproblems. Chang \emph{et al.} \cite{Chang2011Ordinal} proposed
an ordinal hyperplanes ranker to employ the information of relative order between ages. Niu \emph{et al.}
\cite{Niu2016Ordinal} addressed the ordinal regression problem with multiple output CNN. Chen \emph{et al.}
\cite{Chen2017Using} presented Ranking-CNN and established a much tighter error bound for ranking based
age estimation. However, the relations between binary subproblems are ignored in these methods, and ordinal
regression is limited to scalar output \cite{Huang2017Soft}.

 \textbf{Random Forests:}
Random forests \cite{Breiman2001Random} is a widely used classifier in  machine learning and  computer
vision community. Their performance has been empirically demonstrated in many tasks such
as human pose estimation \cite{Jamie2013Efficient} or image classification \cite{Bosch2007Image}.
Meanwhile, deep CNN \cite{Krizhevsky2012ImageNet, He2016Deep} shows the superior performance
of feature learning. Deep neural decision forests (dDNFs) were proposed in \cite{kontschieder2015deep} to combine
these two worlds. Each neural decision tree consists of several split nodes and leaf nodes. Each split node decides
the routing direction in a probabilistic way, and each leaf node holds a class-label distribution. The dDNFs are
differentiable, and the split nodes and leaf nodes are alternating learned using a two-step optimization strategy.
As a classifier, dDNFs have shown the superior results on many classification tasks.
There have been some efforts to migrate dDNFs to the regression problem. Shen \emph{et al.}   proposed DRF
for age estimation by extending the distribution of leaf node to the  continuous Gaussian distribution
\cite{Shen_2018_CVPR}. NRF \cite{Roy2016Monocular} was designed for monocular depth estimation, which used CNN layers
to build the structure of random forests.  However, as will be mentioned in Sec. 3, it is not
suitable to use tree architectures directly in some regression tasks, such as age estimation.

\section{Proposed Approach}

\begin{figure*}[t]
\begin{center}
   \includegraphics[width=1.0\linewidth]{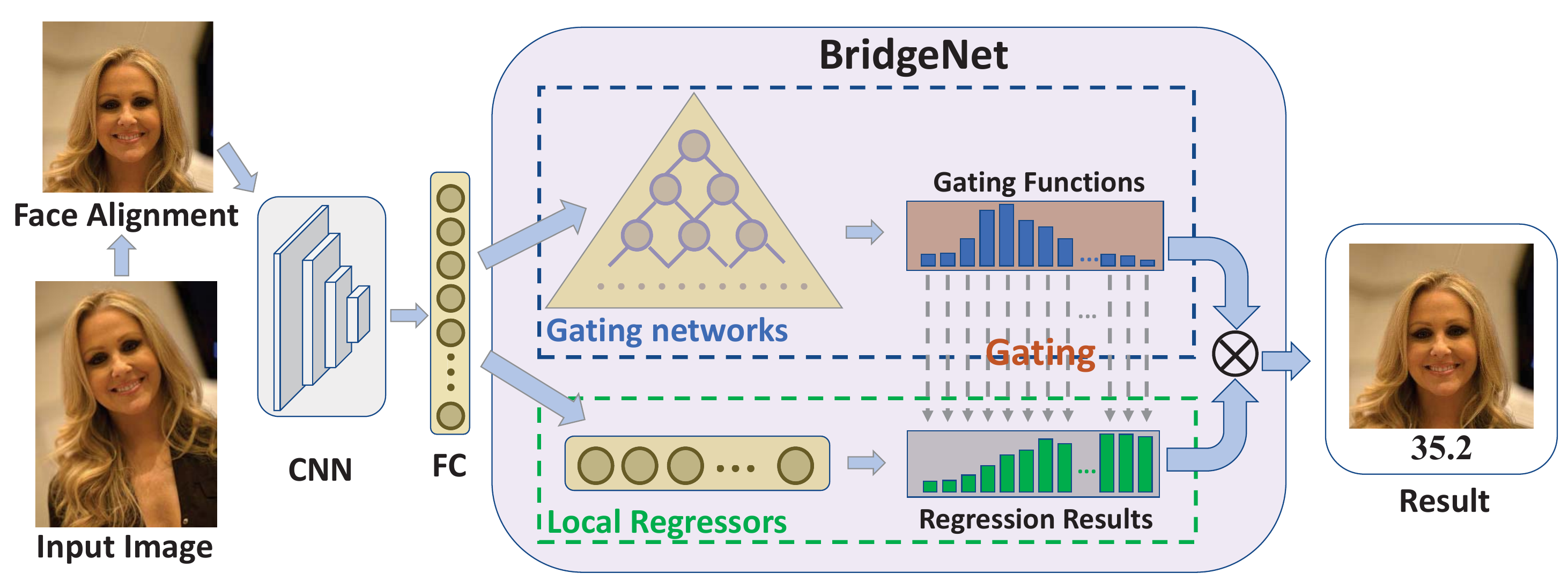}
\end{center}
   \caption{Flowchart of our proposed method for age estimation. For a given input image, we first apply a face
   alignment algorithm to get an aligned facial image. Then the aligned image is passed through a CNN for feature
   extraction. The extracted features are connected with two parts of BridgeNet: local regressors and gating
   networks separately. Gating networks generate continuity-aware gating functions to weight the regression
   results provided by local regressors. The final age is computed by summing the weighted regression results.
   }
\label{fig:flowchart}
\end{figure*}
\subsection{Overall Framework}
The flow chart of our method is illustrated in Fig \ref{fig:flowchart}.
For any input image $\bm{x}  \in \mathcal{X} $, we first crop the human face from the image to remove the background
and then align
the face. The aligned face image is sent to a deep convolution neural network to extract
features. Then the features are connected with two parts of BridgeNet: local regressors
and gating networks separately.  The final age is estimated as a weighted combination over all the
local regressors.

The local regressors are utilized to handle heterogeneous data, which splits the training data
into $k$ overlapping subsets. Each subset is used to learn a local regressor. We denote
$\bm{y} \in \mathcal{Y} $ as the output target of input sample $\bm{x} \in \mathcal{X}$, so the regressor
of the $l^{th}$ subset ($l = 1,2,...,k$) can be formulated as:
\begin{equation}
f(\bm{y}|\bm{x},z = l) = \mathcal{N}(\bm{y}| \mu_l(\bm{x}), \sigma_l^2),
\label{equ:gaussian}
\end{equation}
where $z$ is a latent variable that denotes the affiliation of $\{\bm{x,y}\}$ to a subset,
 and $\mu_l(\bm{x})$  denotes the regression result of the $l^{th}$ local regressor for
input sample $\bm{x}$. Moreover, a Gaussian distribution $\mathcal{N}(\bm{y})$ with a mean
of $\mu_l(\bm{x})$ and a variance of $\sigma_l^2$ is used to model the regression error.


In order to combine these regression results effectively, the gating networks with a new
bridge-tree architecture are proposed, which generate a gating function for each local
regressor. We denote the gating function corresponding to the $l^{th}$ local regressor
as $\pi_l(\bm{x})$. Clearly,
$\pi_l(\bm{x})$s are positive and $\sum_{l} \pi_l(\bm{x}) = 1$ for any $\bm{x} \in \mathcal{X} $.
Then we can address age estimation by modeling the conditional
probability function:
\begin{equation}
p(\bm{y|x}) = \sum_{l} \pi_l(\bm{x}) \mathcal{N}(\bm{y}| \mu_l(\bm{x}), \sigma_l^2).
\label{equ:cpf}
\end{equation}
%

The objective of age estimation is to find a mapping $\bm{g: x \rightarrow y }$. The output
$\bm{\hat{y}}$ is estimated for an input sample $\bm{x}$ by calculating the expectation of
conditional probability distribution:
\begin{equation}
\begin{aligned}
\hat{\bm{y}} &=\mathbb{E}[p(\bm{y|x})]  =\mathbb{E} [ \sum_{l} \pi_l(\bm{x}) \mathcal{N}(\bm{y}| \mu_l(\bm{x}), \sigma_l^2) ] \\
&= \sum_{l} \pi_l(\bm{x}) \mu_l(\bm{x}).
\end{aligned}
\label{equ:regress}
\end{equation}

So the sum of regression results weighted by gating functions gives the final estimated age.
In the following sections, we will provide a detailed description of how local regressors
 and gating networks generate regression results and continuity-aware gating functions respectively.

\subsection{Local Regressors}
As a divide-and-conquer approach, local regressors can be used to model heterogeneous
data effectively. Local regressors divide the data space into multiple subspaces, and
each local regressor only performs regression on one subspace.
We can regard local regressors as multiple experts.
Each expert has good knowledge in a small regression region, and different experts
cover different regression regions. So the ensemble of experts can give a desirable result
even with heterogeneous data.

Here, we divide data by age labels, and each regressor is assigned data in an age group.
The mediums of the regression regions of local regressors are evenly distributed throughout the
whole regression space, and all local regressors have the same length of regression region.

To further model the continuity of age labels, we let the regression regions of local regressors
are densely overlapped. The adjacent local regressors have a very high overlap in their responsible
regions, which makes them have a high similarity. Therefore, for any value, there are multiple regressors
responsible for regressing it, which allows us to employ ensemble learning to make the regression result
 more accurate.


\begin{figure}[t]
  \centering

  \subfigure[Illustration how to build a four-layer binary bridge-tree.
  Node $o_5$ and  $o_6$, node $l_2$ and  $l_3$, node $l_6$ and $l_7$
  in the binary tree are merged into node $o_5$, $l_2$ and $l_3$ in the binary
  bridge-tree respectively. Node $l_4$ and $l_5$ are truncated.
   ]{
    \label{fig:pyramidRoute:a}
    \includegraphics[width=1.0\linewidth]{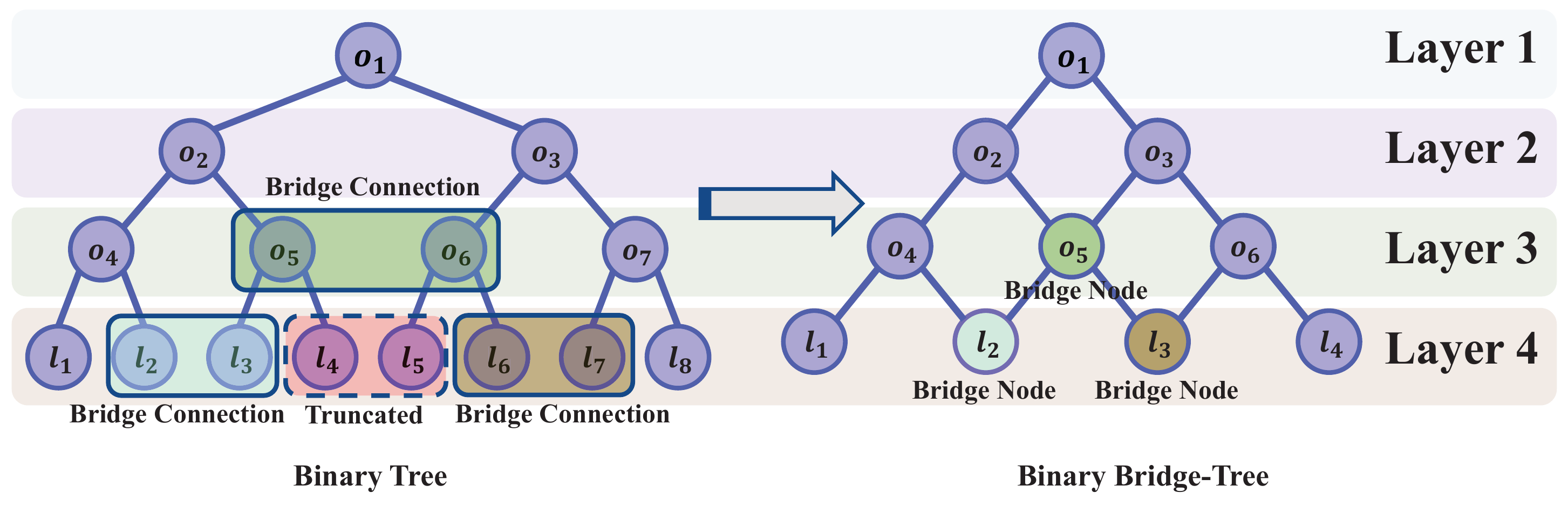}
  }

  \subfigure[Illustration how to build a three-layer triple bridge-tree.
  Node $l_3$ and  $l_4$, node $l_6$ and  $l_7$
  in the triple tree are merged into node $l_3$ and $l_5$  in the triple
  bridge-tree respectively.
  ]{
    \label{fig:pyramidRoute:b}
    \includegraphics[width=1.0\linewidth]{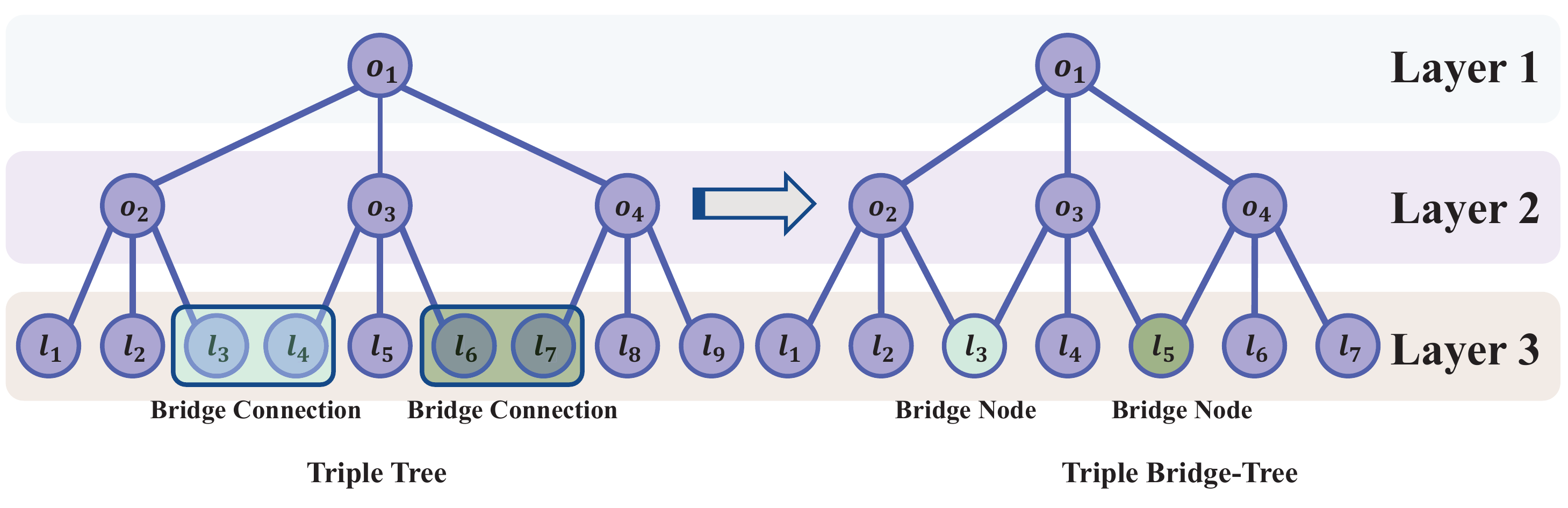}
  }
  \caption{Illustration how to build a bridge-tree}
  \label{fig:pyramidRoute}
\end{figure}

\subsection{Gating Networks}

\textbf{Bridge Connections:}  The design of local regressors follows the principle of divide-and-conquer.
In our approach, gating networks are required to decide the weights of local regressors. Therefore, using
gating networks with a divide-and-conquer architecture makes the gating networks and local regressors
better cooperate with each other. The tree structure is a widely used hierarchical architecture with the
divide-and-conquer principle. For example, the decision tree is a popular classifier in  machine learning
and  computer vision community, which has a tree structure and a coarse-to-fine decision-making
process.

On the other hand, there is a continuity relationship between local regressors due to the continuous aging
process. The design of densely overlapped local regressors further strengthens this relationship.
However, directly using tree structure can not well model this relationship between local regressors,
considering that the leaves of the decision tree are independent class labels, while the leaves of our method
are local regressors with a strong relationship.
For example, the leaf node $l_4$ and $l_5$ in the left side of Figure
\ref{fig:pyramidRoute:a} are adjacent leaf nodes, but their first common ancestor node is the root node,
so the similarity between $l_4$ and $l_5$ caused by continuity can't  be well modeled.

We introduce bridge connections into tree models to enforce the similarity between neighbor nodes.
For two adjacent nodes on the same layer, the rightmost child of the left node and the leftmost child of the right
node are merged into one node. We call this operation a bridge connection because it connects two distant
nodes like a bridge. The merged point, which is named bridge node here, plays a role in
communicating information between the child nodes of the left node and the child nodes of the right node.
By applying this operation to  a tree model layer by layer, a new continuity-aware structure  named bridge-tree
is obtained.

Figure \ref{fig:pyramidRoute:a} shows how to get a 4-layer binary bridge-tree by applying bridge connections to
a 4-layer binary tree.
We can see in the binary bridge-tree that the rightmost child of node $o_2$ and the leftmost
child of node $o_3$ are merged into node $o_5$. Bridge node $o_5$ is the information communication bridge between
the child nodes of node $o_2$ and the child nodes of node $o_3$.
The same operation is applied to node $l_2$ and  $l_3$, node $l_6$ and
$l_7$ in the binary tree. They are merged into node $l_2$ and $l_3$ in binary
bridge-tree respectively.  Node $l_4$ and  $l_5$ in binary tree are
truncated because that node $o_5$ and  $o_6$ in the binary tree have already been merged into one node.
Furthermore, the bridge connection can be
applied to  multiway tree to get  multiway bridge-tree. Especially, Figure \ref{fig:pyramidRoute:b}
gives another example of how to build a triple bridge-tree.
It is worth noting that the growth rate of
node number of the triple bridge-tree is very close to that of the binary tree.

\textbf{Gating Functions:}  In this section, we will describe how to use bridge-tree
structured gating networks to generate continuity-aware gating functions.
 Bridge-tree contains two types of nodes: decision (or split) nodes
and prediction (or leaf) nodes. The decision nodes indexed by $\mathcal{O}$ are internal
nodes, and the prediction nodes indexed by $\mathcal{L}$ are the terminal nodes.
Each prediction node $l \in \mathcal{L}$ corresponds to a regression result $\mu_l(\bm{x})$ and
a gating function $\pi_l(\bm{x})$. The regression results are given by local regressors while the
gating functions are given by gating networks.

To facilitate the later parts of this paper,
$\mathcal{N}$ is used to index all nodes in bridge-tree and $\mathcal{E}$
is used to index all edges in bridge-tree.  We also denote $F_n$ and $C_n$ as the parent
nodes set and the child nodes set of node $n \in \mathcal{N}$, respectively.
When a sample $\bm{x} \in \mathcal{X}$ reaches a decision node $o$, it will be sent to the
children of this node. Following \cite{kontschieder2015deep,Shen_2018_CVPR,NIPS2017_6685},
we use a probabilistic soft decision. Every edge $e \in \mathcal{E}$ is attached with
a probability value. The edges connecting decision node $o$ and its child nodes
form a decision probability  distribution at node $o$. So that means $e_o^m(\bm{x})$s are
positive for any node $m \in C_o$ and $\sum_{m \in C_o} e_o^m(\bm{x}) = 1$,
%
where $e_o^m(\bm{x})$ represents the probability value which sits at the edge from node $o$
to node $m$. Once a sample ends in a leaf node $l$, the gating function for leaf
node $l$ can be
obtained by accumulating all the probability values of the path from the root node to the
leaf node $l$. For example, there are three paths from root node $o_1$ to leaf node $l_2$ in
the binary bridge-tree in Figure \ref{fig:pyramidRoute:a}: $o_1-o_2-o_4-l_2$,  $o_1-o_2-o_5-l_2$,
and  $o_1-o_3-o_5-l_2$. So the gating function for leaf node $l_2$ can be computed as
$\pi_{l_2}(\bm{x}) =  e_{o_1}^{o_2}(\bm{x}) e_{o_2}^{o_4}(\bm{x}) e_{o_4}^{l_2}(\bm{x}) +
 e_{o_1}^{o_2}(\bm{x}) e_{o_2}^{o_5}(\bm{x}) e_{o_5}^{l_2}(\bm{x}) +
  e_{o_1}^{o_3}(\bm{x}) e_{o_3}^{o_5}(\bm{x}) e_{o_5}^{l_2}(\bm{x})$.
Moreover, we give a recursive expression of gating function
by extending the definition of gating function to all nodes $n \in \mathcal{N}$:
\begin{equation}
\pi_{n_0}(\bm{x}) = 1
\label{equ:gating:1}
\end{equation}
\begin{equation}
\pi_n(\bm{x}) = \sum_{m \in F_n} \pi_m(\bm{x}) e_m^n(\bm{x}),
\label{equ:gating:2}
\end{equation}
where $\pi_n(\bm{x})$  denotes the gating function for node $n$
and node $n_0$ is the root node of bridge-tree. We establish a one-to-one correspondence between
the gate networks and the probability values on the edges of  bridge-tree, that is to say,
every gating network corresponds to a probability value which sits at an edge of the bridge-tree.
Then the gating functions for leaf nodes can be calculated using the outputs of gating networks
in the above recursive way.


\begin{figure}[t]
\begin{center}
   \includegraphics[width=1.0\linewidth]{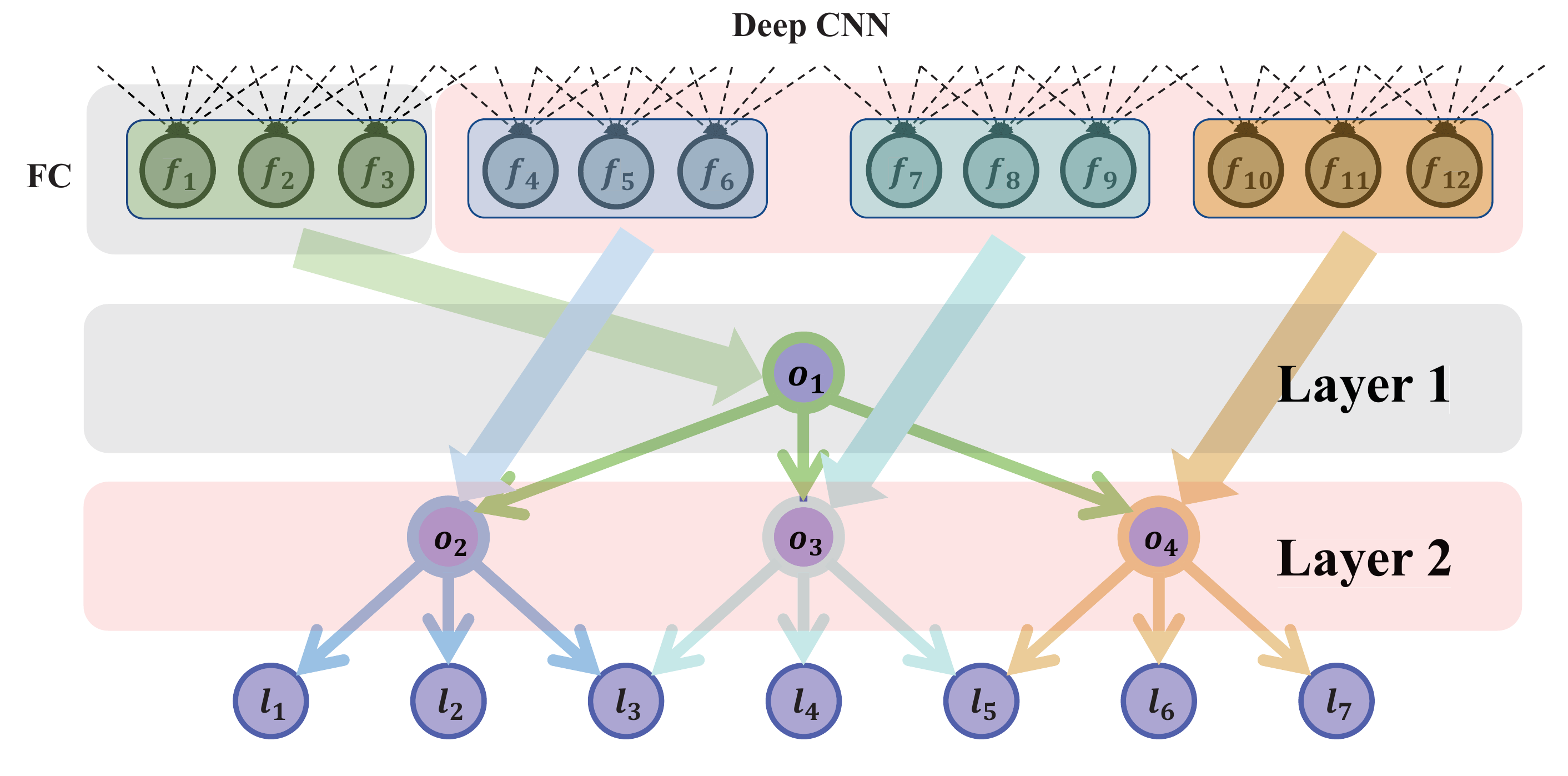}
\end{center}
   \caption{Illustration how to implement gating networks. An FC layer connected with deep CNN
   is employed. Each neuron in the fully-connected layer corresponds to an edge of bridge-tree.
   For example, neural $f_1$, $f_2$ and $f_3$ correspond to edge $o_1$-$o_2$, $o_1$-$o_3$, and
   $o_1$-$o_4$ respectively. For the triple bridge-tree, every three neurons are normalized using
   a softmax layer. Then the normalized outputs of neurons give all the probability values
   on the edges of bridge-tree. Finally, the gating functions for leaf nodes are calculated using
   Eq. \ref{equ:gating:1} and Eq. \ref{equ:gating:2}.
}
\label{fig:implegating}
\end{figure}

\subsection{Implementation Details}
We employ  a fully-connected layer to implement densely overlapped local regressors. The sigmoid function
is utilized as the activation function. Then each local regressor maps the activation value to their
regression space as the expert result. As mentioned above, we use $\mu_l(\bm{x})$ to denote the result
of the $l^{th}$ local regressor , then the regression loss is given by:
\begin{equation}
L_{reg}(\bm{x,y}) =  \sum_{l \in \mathcal{L}} {\mathbb I}_l(\bm{x,y}) (\bm{y} - \mu_l(\bm{x}))^2 ,
\label{equ:loss_expert}
\end{equation}
where ${\mathbb I}_l(\bm{x,y})$ denotes if  $\bm{y}$ is located in the responsible region of the $l^{th}$local
regressor.

Figure \ref{fig:implegating} demonstrates the implementation of gating networks, which
also employs a fully-connected layer. Each neuron in the
fully-connected layer corresponds to an edge of bridge-tree. We let $B$ represents
the number of branches of each decision node. Considering that $B$ edges starting
from the same node form a probability distribution, we apply a softmax function
to every $B$  neurons of the fully connected layer for normalization. The gating functions
of leaf nodes can be calculated using these normalized outputs of neurons according to Eq.
\ref{equ:gating:1} and Eq. \ref{equ:gating:2}.

Since the ground truth for supervising gating functions is not available,
we  build approximated gating targets for an input sample $(\bm{x,y})$ as follow:
\begin{equation}
\hat{\pi}_l(\bm{x})=\frac{1}{R}{\mathbb I}_l(\bm{x,y}),
\label{equ:target}
\end{equation}
where $R = \sum_{l}{\mathbb I}(\bm{x,y})$ is used for normalization.
Although the labels are not accurate, our gating networks can be aware of the continuity
between local regressors, so a satisfying result can be achieved even with weakly supervised
signals.


The KL divergence is utilized as the loss term to train the gating networks of BridgeNet:
\begin{equation}
L_{gate}(\bm{x,y}) = - \sum_{l \in \mathcal{L}} \hat{\pi}_l(\bm{x}) \log(\pi_l(\bm{x})).
\label{equ:loss:gating}
\end{equation}

In the end, we  jointly learn local regressors and gating networks by defining the total
loss as follow:
\begin{equation}
L_{total}(\bm{x,y}) = L_{reg}(\bm{x,y}) + \lambda L_{gate}(\bm{x,y}),
\label{equ:loss:total}
\end{equation}
where $\lambda$ is used to balance the importance between the regression task and gating task.

We  observe that the proposed BridgeNet can be easily implemented by using typically available fully-connected,
softmax and sigmoid layers in the existing deep learning frameworks such as TensorFlow
\cite{Abadi2016TensorFlow}, PyTorch \cite{paszke2017automatic}, etc.
Furthermore, our fully differentiable BridgeNet
can be embedded within any deep convolutional neural networks, which enables us to conduct end-to-end
training and obtain a better feature representation.

\section{Experiments}
In this section, we first introduce the datasets and present some details about our experiment settings.
Then we demonstrate the experimental results to show the effectiveness of the proposed BridgeNet.

\subsection{Datasets}

\textbf {MORPH II} is the largest publicly available longitudinal face dataset and the most
popular dataset for age estimation. This dataset includes more than 55,000 images from about 13,000
subjects and age ranges from 16 to 77 years.

In this paper, two widely used protocols are employed for evaluation on MORPH II. The first setting
uses a subset of MORPH II as described in
\cite{Chang2011Ordinal,Chen2013Cumulative,Wang2015Deeply}. This setting selects 5,492 images of people
of Caucasian descent to avoid the cross-race influence. Then these 5,492 images are randomly divided
into two non-overlapped parts: 80\% of data for training and 20\% of data for testing.
The second
setting  used in \cite{Yi2014Age,Guo2011Simultaneous} randomly splits the whole MORPH II dataset
into three non-overlapped subsets $S1, S2, S3$ following the rules detailed in \cite{Yi2014Age}. The training
and testing are repeated twice in this setting: 1) training on $S1$, testing on $S2 + S3$ and 2) training
on $S2$, testing on $S1 + S3$. We will report the performance of these two experiments and their average.

\textbf {FG-NET} consists of 1002 color or greyscale face images of 82 individuals with ages ranging from
0 to 69 years old subjects. For evaluation, we adopt the setup of \cite{Guo2008Image,Rothe2016Deep},
which uses leave-one person-out (LOPO) cross-validation. The average performance over 82 splits is reported.

\textbf {Chalearn LAP 2015}  is the first dataset on apparent age estimation. For any image, at least
10 independent users are required to give their opinions and then the average age is used as the annotation.
Additionally, the standard deviation of opinions for a given image is also provided. This dataset contains 4699
images, where 2476 images for training, 1136 images for validation, and 1087 images for testing. The age range is from
0 to 100 years old.

\textbf {IMDB-WIKI}  contains more than half a million labeled images of celebrities, which are crawled
from IMDb and Wikipedia. This datatset contains too much noise, so it is not suitable for evaluation.
However, it is still a good choice to use this dataset for pretraining after data cleaning. We select about 200
thousand images according to the setting in \cite{Rothe2016Deep} to pre-train our network.

\begin{table}
\caption{The comparisons between the proposed method and other state-of-the-art methods on MORPH
II dataset (setting I) and FG-NET dataset. }
\begin{center}
\begin{tabular}{|l|c|c|c|c|}
\hline
Method  &  MORPH II  &FG-NET & Year \\
\hline\hline
Human \cite{Han2015Demographic} & 6.30 &  4.70 & - \\
AGES \cite{Geng2007Automatic} & 8.83 & 6.77 & 2007 \\
IIS-LDL \cite{Geng2010Facial} & -  & 5.77 & 2010 \\
CPNN \cite{Geng2013Facial} & -  & 4.76 & 2013  \\
MTWGP \cite{Zhang2010Multi} & 6.28 & 4.83  & 2010\\
OHRank \cite{Chang2011Ordinal} & 6.07   & 4.48  & 2011\\
CA-SVR \cite{Chen2013Cumulative} & 5.88   & 4.67 & 2013 \\
DRFs \cite{Shen_2018_CVPR} &  2.91  & 3.85 &2018  \\
DEX \cite{Rothe2016Deep}  &  2.68    & 3.09 & 2016 \\
Pan \emph{et al.} \cite{Pan_2018_CVPR} & -   & 2.68  & 2018 \\
\hline
\textbf{BridgeNet}  & \textbf{2.38} & \textbf{2.56} & -\\
\hline
\end{tabular}
\end{center}
\label{table:SOTA:MORPH1}
\vspace {-0.3cm}
\end{table}

\begin{table}
\caption{The results on MORPH II dataset (setting II).
The performance of two different settings and their average
are reported. Our method achieves the state-of-the-art
performance.}
\begin{center}
\begin{tabular}{|l|c|c|c|c|}
\hline
Method & Train & Test  & MAE & Avg  \\
\hline  \hline
\multirow{2}*{KPLS \cite{Guo2011Simultaneous}} & S1 & S2+S3 & 4.21 & \multirow{2}*{4.18} \\
~ & S2 & S1+S3 & 4.15 & ~ \\
\hline
\multirow{2}*{BIF+KCCA \cite{Guo2013Joint}} & S1 & S2+S3 & 4.00 & \multirow{2}*{3.98} \\
~ & S2 & S1+S3 &3.95 & ~ \\
\hline

\multirow{2}*{CPLF \cite{Yi2014Age}} & S1 & S2+S3 & 3.72 & \multirow{2}*{3.63} \\
~ & S2 & S1+S3 & 3.54 & ~ \\
\hline
\multirow{2}*{Tan \emph{et al.} \cite{ACCV2016Age}} & S1 & S2+S3 & 3.14 & \multirow{2}*{3.03} \\
~ & S2 & S1+S3 & 2.92 & ~ \\
\hline
\multirow{2}*{DRFs \cite{Shen_2018_CVPR}} & S1 & S2+S3 & - & \multirow{2}*{2.98} \\
~ & S2 & S1+S3 & - & ~ \\
\hline
\multirow{2}*{\textbf{BridgeNet}} & S1 & S2+S3 & \textbf{2.74} & \multirow{2}*{\textbf{2.63}} \\
~ & S2 & S1+S3 & \textbf{2.51} & ~ \\
\hline
\end{tabular}
\end{center}
\label{table:SOTA:MORPH2}
\vspace {-0.5cm}
\end{table}

\begin{table*}
\caption{ Comparisons with the state-of-the-art
 methods on the Chalearn LAP 2015 dataset
 }
\vspace {0.3cm}
\begin{center}
\begin{tabular}{|c|c|cc|cc|c|c|c|}
\hline
\multirow{2}*{Rank} & \multirow{2}*{Team} & \multicolumn{2}{c|}{Validation Set} & \multicolumn{2}{c|}{Test Set} & Pretrain & \multirow{2}*{Netwrok} & \# of \\
~ & ~ & MAE & $\epsilon$-error & MAE & $\epsilon$-error & Set & ~ & Networks \\
\hline\hline
- & \textbf{BridgeNet} & \textbf{2.98} & \textbf{0.26} & \textbf{2.87} & \textbf{0.255140} & IMDB-WIKI & VGG-16 &  \textbf{1} \\
- & Tan \emph{et al.} \cite{Tan2017Efficient} & 3.21 & 0.28 & 2.94 & 0.263547 & IMDB-WIKI & VGG-16 & 8 \\
\hline
1 & CVL\_ETHZ \cite{Rothe2016Deep} & 3.25 & 0.28 & - & 0.264975  & IMDB-WIKI & VGG-16 & 20 \\
2 & ICT-VIPL \cite{Liu2015AgeNet} &  3.33 & 0.29 & - & 0.270685 & MORPH, CACD, \emph{et al.} & GoogleNet &  8 \\
3 & WVU\_CVL \cite{Zhu2015A}  & - & 0.31 & - & 0.294835 &MORPH, CACD, \emph{et al.} &  GoogleNet & 5 \\
4 & SEU\_NJU \cite{Yang2015Deep} & - & 0.34 & - &  0.305763 & FG-NET,MORPH,\emph{et al.}& GoogleNet & 6 \\
~ & Human & - & - & - & 0.34 & - & - & - \\
\hline
\end{tabular}
\end{center}
\label{table:SOTA:LAP}
\vspace {-0.5cm}
\end{table*}

\subsection{Experimental Settings}
Face alignment is a common preprocessing step for age estimation. First, all images are sent to MTCNN
\cite{Zhang2016Joint} for face detection. Then we align all the face images by similarity
transformation based on the detected five facial
landmarks. After that, all images are resized into $256 \times 256$.

Data augmentation is an effective way to avoid overfitting and improve the generalization of deep networks,
especially when the training data is insufficient. Here, we augment training images with horizontal
flipping and random cropping.

VGG-16 \cite{Simonyan15} is employed as the basic backbone network of the proposed method. We first initialize the VGG-16 network
with the weights from training on ImageNet 2012 \cite{Olga2015ImageNet} dataset. Then the network is pre-trained
on IMDB-WIKI dataset.
To optimize the proposed network, we use the mini-batch stochastic gradient descent (SGD) with batch size 64
and apply the Adam optimizer \cite{Kingma2014Adam}.
The initial learning rate is set to 0.0001 for experiments on MORPH II dataset.
The training images on FG-NET and Chalearn LAP 2015 datasets are extremely insufficient, so we set the initial
learning rate of CNN part to 0.00001 for the experiments on these datasets to avoid overfitting. The initial
learning rate of the BridgeNet part is still 0.0001 on these datasets to accelerate convergence. We train our network
for 60 epochs
and set $\lambda$ to 0.001 to balance the gating loss and regression loss.  The length of regression region for local
regressors is set to 25. We choose a triple bridge-tree with a depth of 5 as the architecture of our BridgeNet, which
is a trade-off of efficiency and complexity.
Our algorithm is implemented within the PyTorch \cite{paszke2017automatic} framework.
A GeForce GTX 1080Ti GPU is used for neural network acceleration.

\subsection{Evaluation Metrics}
The mean absolute error (MAE) and cumulative score (CS) are used as evaluation metrics on MORPH II and FG-NET
datasets. MAE is calculated using the mean absolute errors between the estimated result and ground truth:
$ MAE =  \frac{1}{K} \sum_{i=1}^{K} |\bm{y_i'} - \bm{y_i}|$,
 where $\bm{y_i'}$ denotes the predicted age value for the $i^{th}$ image, and $K$ is the number of testing samples.
 Obviously, a lower MAE result means better performance.
 $CS(\theta)$ is computed as follows:
 $CS(\theta) = \frac{K_{\theta}}{K}$,
where $K_{\theta}$ represents the number of test images whose absolute error between the estimated result and the ground
truth is not greater than $\theta$ years.
Naturally, the higher the $CS(\theta)$, the better performance it gets.
The $\epsilon$-error was proposed by the Chalearn LAP challenge as a quantitative measure,
which is defined as: $\epsilon = 1 - \frac{1}{K} \sum_{i=1}^K e^{- \frac{(\bm{y_i' - y_i} )^2}{2 \sigma_i^2}}$,
where $\sigma_i$ is the standard deviation of the $i^{th}$ image. Clearly, a lower $\epsilon$-error means better
performance.

\subsection{Results and Analysis}

\begin{figure*}[t]
  \centering
  \subfigure[MORPH II (setting I)]{
    \label{fig:cs:morph1}
    \includegraphics[width=0.32\linewidth]{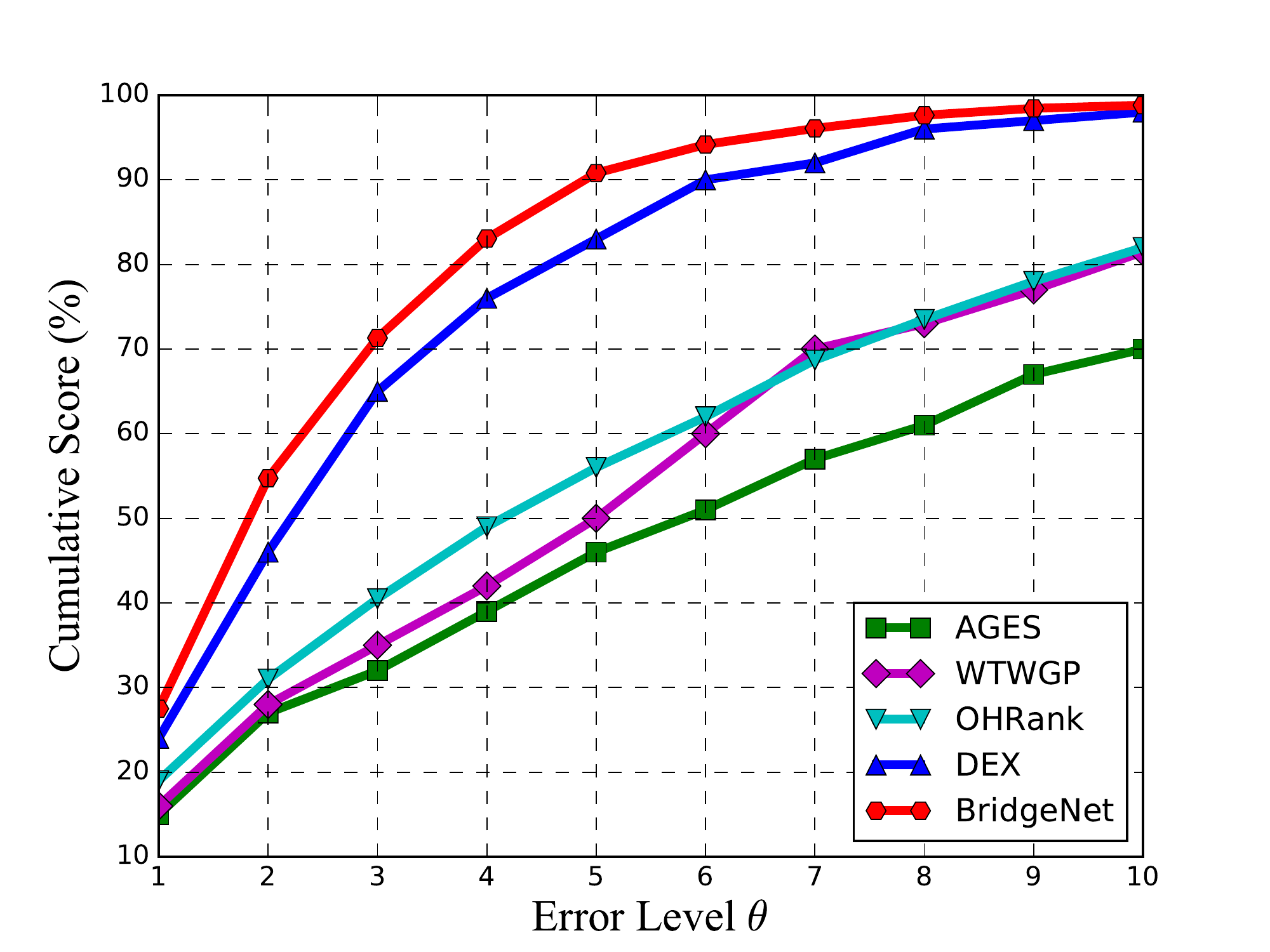}
  }
  \subfigure[MORPH II (setting II)]{
    \label{fig:cs:morph2}
    \includegraphics[width=0.32\linewidth]{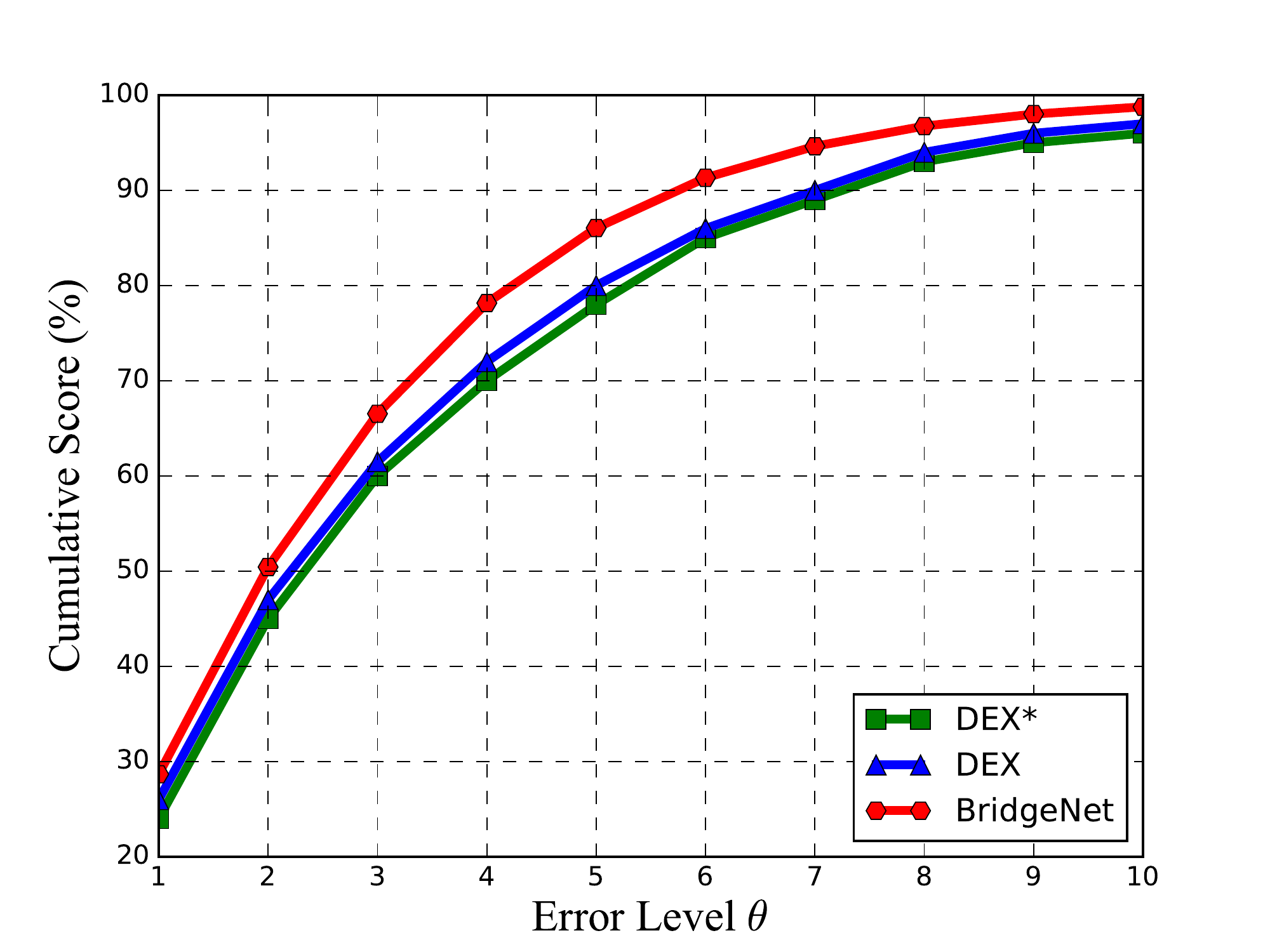}
  }
  \subfigure[FG-NET]{
    \label{fig:cs:fegnet}
    \includegraphics[width=0.32\linewidth]{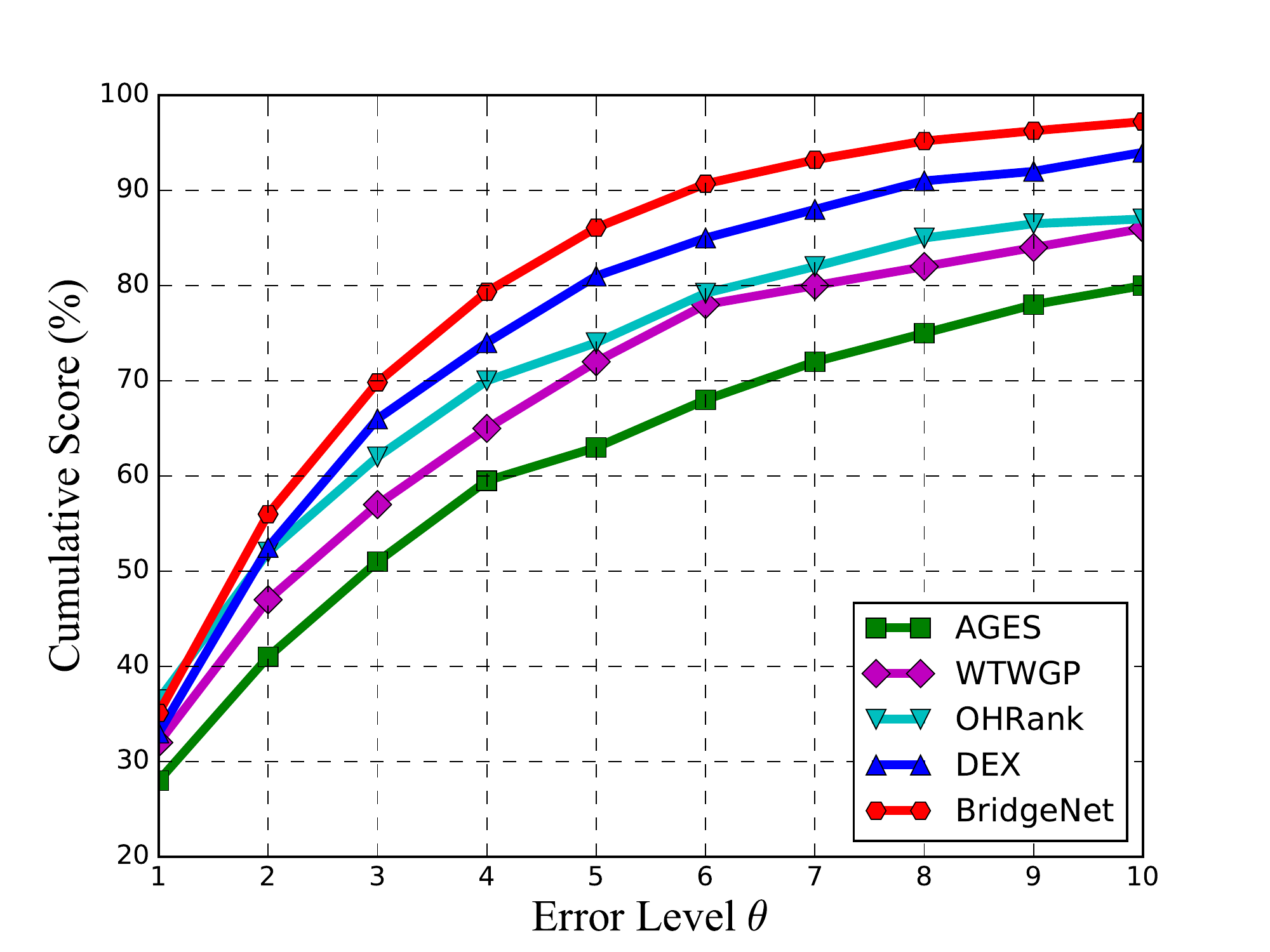}
  }
  \caption{(a) CS curves compared with other methods on MORPH II dataset with
  setting I. (b)CS curves compared with other methods on MORPH II dataset with
  setting II. * means that the IMDB-WIKI dataset was not used to pre-train the model.
  (c) CS curves compared with other methods on FG-NET.}
  \label{fig:cs}
\end{figure*}

\begin{table*}
\caption{The comparisons of different architectures on MORPH II dataset (setting I)}
\vspace {0.3cm}
\begin{center}
\begin{tabular}{|l|cccc|cccc|cccc|}
\hline
Architecture & \multicolumn{4}{c|}{Softmax} & \multicolumn{4}{c|}{Tree(binary)} & \multicolumn{4}{c|}{Bridge-Tree(triple)} \\
\hline \hline
Depth & \multicolumn{4}{c|}{-} & 4 & 5 & 6 & 7 & 3 & 4 & 5 & 6 \\
Num. of leaf nodes & 16 & 32 & 64 & 128 & 16 & 32 & 64 & 128 & 15 & 31 & 63 & 127 \\
Num. of decision nodes & \multicolumn{4}{c|}{-} & 15 & 31 & 63 & 127 & 11 & 26 & 57 & 120 \\
MAE & 2.68 & 2.59 & 2.54 & 2.53 & 2.66 & 2.54 & 2.51 & 2.49 & \textbf{2.51} & \textbf{2.43} & \textbf{2.38} & \textbf{2.38} \\
\hline
\end{tabular}
\end{center}
\label{table:Alba:triple}
\vspace {-0.7cm}
\end{table*}

\begin{table}
\caption{The results of binary bridge-tree structure on MORPH II dataset (setting I)}
\begin{center}
\begin{tabular}{|l|c|c|c|c|}
\hline
Num. of leaf nodes & 16 & 32 & 64 & 128 \\
\hline
MAE &  2.43 & 2.39 & 2.36 & 2.35   \\
\hline
\end{tabular}
\end{center}
\label{table:Alba:binary}
\vspace {-0.8cm}
\end{table}

\textbf{Comparisons with the State-of-the-art:}
We first compare the proposed BridgeNet with other state-of-the-art methods on MORPH  II dataset
with different settings and FG-NET dataset. Table \ref{table:SOTA:MORPH1} and Table
\ref{table:SOTA:MORPH2} show the results on MORPH II and
FG-NET using MAE metric. The results demonstrate that our method outperforms the state-of-the-art methods with
a clear margin on both datasets.
 Our method achieves the lowest MAE of 2.38,
2.63, and 2.56 on MORPH II with setting I, MORPH II with setting II, and FG-NET respectively.
The classification based methods, such as
DEX \cite{Rothe2016Deep}, Tan \emph{et al.} \cite{ACCV2016Age}, are not optimal because they treat different ages
as independent class labels. On the other hand, the ranking based methods, such as
OHRank \cite{Chang2011Ordinal}, can't capture the continuity relationship among components, resulting in
unsatisfactory performance.
DRFs \cite{Shen_2018_CVPR} uses a tree structure to weight several Gaussian
distributions and Pan \emph{et al.} propose a mean-variance loss for age estimation.
Both of them can't effectively model the continuous property of the aging process.
The
CS comparisons with the state-of-the-art methods on MORPH II and FG-NET are shown in Figure \ref{fig:cs}.
The experimental results show that our approach consistently outperforms other methods.

In addition to these two datasets, we present results of our method on Chalearn LAP 2015 dataset. Following
\cite{Rothe2016Deep,Rothe-ICCVW-2015,Tan2017Efficient}, a few tricks are used on this competition dataset.
To get the  performance on the test set, we finetune our network on both training and validation sets after
finetuning on IMDB-WIKI dataset. In the test phase, for any given image, we crop it into four corners and a
central crop, then the five crops plus the flipped version of these are sent to our network, and  these
ten predictions are averaged. It is important to note that we only use these tricks on Chalearn LAP 2015
dataset. To make a more comprehensive comparison, we also show the performance on the validation set, which
only uses the training set to finetune. The experimental results are shown in Table \ref{table:SOTA:LAP}.
The  bottom half of
the table shows the results of the participating teams, and the top half shows the results of our method and
another state-of-the-art method. We can see that our method achieves  better performance than other methods.
Our method achieves an MAE of 2.98 and a $\epsilon$-error of 0.26 on the validation set, which reduces the
state-of-the-art performance by 0.23 years for MAE and 0.02 for $\epsilon$-error. For the test set,
we also achieve a lower MAE and $\epsilon$-error. All of the above results of our method are obtained by
using a single network, while others methods use an ensemble of multiple networks, which further
illustrates the superiority of our method.

\textbf{Ablation Study and Parameters Discussion:}
To validate the effectiveness of the proposed BridgeNet, we compare it with two  baseline
architectures: one uses a tree structure to construct gating functions, and the other uses a softmax
layer to construct gating functions.  To be fair, we use triple bridge-tree structure,
whose node growth rate is close to that
of the binary tree. The experiments are conducted on MORPH II dataset (setting I) and Table
\ref{table:Alba:triple} shows the results.


Several conclusions are drawn from Table \ref{table:Alba:triple}. First, in any of the above architectures, a lower MAE can be
obtained by using more leaf nodes, which is reasonable because more leaf nodes mean more local regressors
and more local regressors mean more expert intelligence. Furthermore, when the number of leaf nodes is large
enough, the performance tends to be saturated. This is because too many leaf nodes make some adjacent
local regressors correspond to the same training data, so it can not increase the actual number of experts.
Second, we  observe that the tree-based method slightly outperforms the softmax based method when the
number of leaf nodes is the same. The tree-based method has a coarse-to-fine, top-to-down decision-making
process as a hierarchical architecture, so it can give better performance than softmax based method. However,
it doesn't explicitly model the continuity relationship between local regressors,
so the performance gain is tiny. Third, bridge-tree based method
(BridgeNet) significantly outperforms the tree-based method at a similar number of leaf nodes even with a shallower
depth and fewer decision nodes. The five layers triple bridge-tree achieves an MAE of 2.38, which reduces the MAE
by 0.13 years compared with the six layers binary tree.
This shows the benefit of introducing bridge connections  and  explicitly
modeling the continuity relation.

To further demonstrate the superiority of bridge-tree, we show the results using binary bridge-tree
architecture on MORPH II dataset (setting I) in Table \ref{table:Alba:binary}. We can see that binary bridge-tree further
improves the accuracy.
This is because, with the same number of leaf nodes, binary bridge-tree
has more bridge nodes, which makes it better capture the continuity relationship between local regressors.

\section{Conclusions}
In this paper, we have presented BridgeNet, a continuity-aware probabilistic network for age estimation.
BridgeNet explicitly models the continuity relationship between different components constructed by
local regressors  using a probabilistic network with a bridge-tree architecture.
Experiments on three datasets
demonstrate that our method is more accurate than other state-of-the-art
methods. Although our method is designed for age estimation, it can also be used for other regression-based
computer vision tasks. In the future work, we plan to investigate the effectiveness of BridgeNet in crowd
counting, pose estimation and other regression-based tasks.

\section*{Acknowledgement}
This work was supported in part by the National Key Research and Development Program of China under Grant 2017YFA0700802, in part by the National Natural Science Foundation of China under Grant 61822603, Grant U1813218, Grant U1713214, Grant 61672306, Grant 61572271, and in part by the Shenzhen Fundamental Research Fund (Subject Arrangement) under Grant JCYJ20170412170602564.

{\small
\bibliographystyle{ieee}
\bibliography{egbib}
}

\end{document}